\documentclass[11pt]{article}
\usepackage[preprint]{acl}
\usepackage{times}
\usepackage{latexsym}
\usepackage[T1]{fontenc}
\usepackage[utf8]{inputenc}
\usepackage{microtype}
\usepackage{inconsolata}
\usepackage{graphicx}
\usepackage{booktabs}
\usepackage{amsmath}
\usepackage{amssymb}
\usepackage{multirow}
\usepackage{xcolor}
\usepackage{orcidlink}

\newcommand{\figref}[1]{\hyperref[#1]{Figure~\ref*{#1}}}
\newcommand{\tabref}[1]{\hyperref[#1]{Table~\ref*{#1}}}
\newcommand{\appref}[1]{\hyperref[#1]{Appendix~\ref*{#1}}}
\newcommand{\secref}[1]{\hyperref[#1]{\S\ref*{#1}}}

\title{A Fixed-Budget, Cluster-Aware Standard for LLM-as-a-Judge Evaluation:\\A Multi-Hop RAG Stress Test}

\author{
Camilo Chacón Sartori\orcidlink{0000-0002-8543-9893}\textsuperscript{*}
\quad
José H. García\orcidlink{0000-0002-5752-4759}\\
\normalfont Catalan Institute of Nanoscience and Nanotechnology (ICN2), CSIC and BIST,\\
\normalfont Campus UAB, Bellaterra, Barcelona, Spain\\
\texttt{camilo.chacon@icn2.cat}
\quad
\texttt{josehugo.garcia@icn2.cat}
\normalfont\\[-0.1em]\small Code and materials: \href{https://github.com/camilochs/fixed-budget-llm-judge-rag}{\texttt{github.com/camilochs/fixed-budget-llm-judge-rag}}
\normalfont\\[-0.1em]\small\textsuperscript{*}Corresponding author
}

\begin{document}
\maketitle

\begin{abstract}
Retrieval-augmented generation (RAG) systems are often compared by asking a large language model (LLM) judge which answer is better. For multi-hop RAG, this has become a measurement problem as much as a modeling problem: the same score can reflect retrieval quality, answer length, lexical overlap, or a statistical test that ignores clustered data. We ask what happens when these choices are made explicit.

We propose a minimum measurement standard for LLM-as-a-judge comparisons in RAG. The standard fixes the top-100 candidate pool, evidence budget, answer cap, generator, and prompt; it also requires pre-registered hypotheses, cluster-aware inference, an exact cluster sign-flip check when feasible, and second-judge replication. Clustered benchmarks can overstate progress; the field should adopt this standard. We stress-test it with Genetic Algorithm Decoder for Multi-hop Evidence Composition (GADMEC), an evolutionary evidence selector, on 400 multi-hop questions in computer science/machine learning (CS/ML) and Materials Science. The protocol changes the empirical story. A binomial test makes all four semantic-baseline comparisons look significant; cluster-aware inference leaves only one Bonferroni-significant result. BM25 beats pure semantic GADMEC under the same budget, while a lexical-semantic hybrid recovers in CS/ML and narrows the Materials Science gap.
\end{abstract}

\section{Introduction}

RAG systems are increasingly compared with pairwise LLM-as-a-judge protocols: two answers are shown to a judge model, and the preferred answer is counted as a win. This protocol is attractive because it is simple and cheap to scale. It also hides important choices. A method can look better because it selected better evidence, because it induced longer answers, because it matched lexical cues that dense retrieval missed, or because the statistical test treated clustered examples as independent. In multi-hop RAG, these mechanisms are easy to mix.

This paper asks whether the headline of a RAG comparison survives when those choices are controlled. In our experiment, the answer is mixed. A binomial test would make all four pre-registered semantic-baseline comparisons look significant. Cluster-aware inference leaves only one comparison significant after Bonferroni correction, with two more significant only before correction. The empirical story therefore depends not only on the selector, but also on the measurement protocol used to evaluate it.

We make two claims. First, pairwise LLM-as-a-judge evaluation for multi-hop RAG is more fragile than standard reporting suggests. Second, the field should adopt a minimum measurement standard for cluster-structured LLM-as-a-judge benchmarks. The standard fixes evidence and answer budgets. It separates confirmatory from exploratory analyses through pre-registration and a public deviation log, uses cluster-aware inference, adds an exact sign-flip check when the cluster count permits it, and replicates headline results with a second judge. These components are not individually new. The contribution is to make them operate together in one controlled RAG comparison and show how the conclusions change.

We stress-test the standard with Genetic Algorithm Decoder for Multi-hop Evidence Composition (GADMEC), a Biased Random-Key Genetic Algorithm (BRKGA; \citealp{goncalves2011brkga}) for evidence subset selection. GADMEC is the instrument, not the protagonist. Its evolutionary search is separated from the decoder that enforces budget and diversity constraints, so a random-fitness ablation can test whether the fitness function contributes beyond the decoder machinery.

\paragraph{Contributions.}
\begin{enumerate}
\item A \textbf{fixed-budget evaluation design} for LLM-as-a-judge comparisons in RAG: same top-100 candidate pool, 2000-token evidence budget, 300-token answer cap, and generator settings for all methods.
\item A \textbf{pre-registered analysis protocol}: four Bonferroni-corrected primary hypotheses, an addendum for length matching and ablations, and a deviation log for post-hoc analyses.
\item A \textbf{cross-domain benchmark}: 687 arXiv papers (2024--2026), a 3-level taxonomy, 10 cross-subfield combinations per area, 20 questions per combination, and 200 contrastive multi-hop questions per area.
\item A \textbf{methodological demonstration}: the headline of a multi-hop RAG comparison can change when inference respects the cluster structure. The binomial test would have reported all four primary tests as significant; wild-cluster bootstrap leaves only one Bonferroni-significant result. The same protocol exposes a lexical-vs-semantic axis: BM25 beats pure semantic GADMEC, while a hybrid recovers in CS/ML.
\end{enumerate}

The paper unfolds as follows. Section~2 situates fixed-budget evaluation relative to LLM-as-a-judge evaluation, multi-hop question answering (QA), subset selection, and pre-registration. Sections~3 and~4 define GADMEC, the baselines, and the controlled evaluation protocol. Section~5 reports the pre-registered results, length controls, content-distance diagnostics, and ablations. Sections~6--8 interpret the mechanisms, conclude, and state the limitations.

\section{Related Work}

\paragraph{LLM-as-a-Judge Evaluation for RAG.} Reference-free LLM-as-a-judge evaluation is now common in RAG research. RAGAS formalised the setup for retrieval-augmented generation \cite{es2024ragas}, and recent surveys catalogue its variants \cite{li2025judgment}. The closest concern for this paper is length bias \cite{zheng2023judging,dubois2024lengthcontrolled}. We control both input and output budgets, then ask what signal remains after length matching.

\paragraph{Multi-hop QA benchmarks.} HotpotQA, 2WikiMultiHop, MuSiQue, and GRADE target chained-evidence multi-hop reasoning \cite{lee2025grade}. Our questions have a different shape: \emph{contrastive composition} (``how do $X$ and $Y$ differ on aspect $Z$''). This makes evidence selection budget-sensitive because a good answer must cover several sub-aspects at once.

\paragraph{Combinatorial subset selection in retrieval.} Maximal Marginal Relevance (MMR; \citealp{carbonell1998mmr}) and Determinantal Point Processes (DPPs; \citealp{kulesza2012dpp}) are classical instances. Genetic algorithms have been applied to RAG primarily for adversarial attack and corpus poisoning \cite{cho2024typos}. We use a BRKGA constructively: the candidate pool is the search space, and the selected subset is the evidence plan.

\paragraph{Pre-registration in NLP.} Pre-registration is rare in natural language processing (NLP). We follow social-science practice with a primary registration, timestamped addenda, and a deviation log separating confirmatory and exploratory analyses.

Together, these strands motivate an evaluation instrument rather than only a new retriever. The selector is allowed to vary; the evidence budget, answer cap, judge protocol, and evidential status of each analysis are not.

\begin{figure*}[t]
\centering
\includegraphics[width=\textwidth]{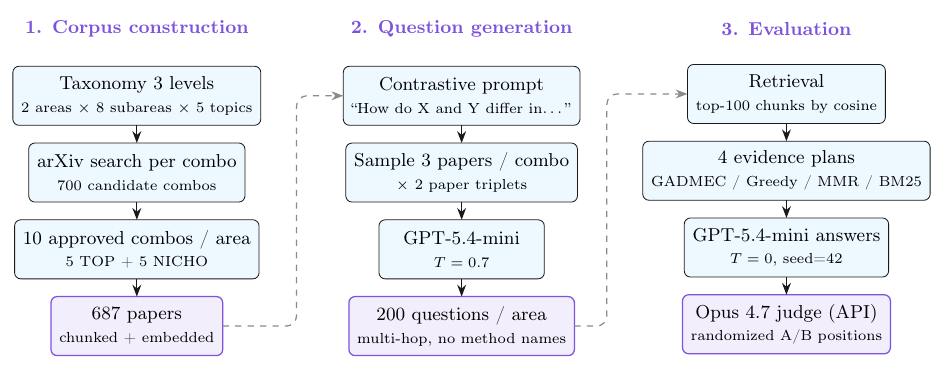}
\caption{GADMEC pipeline. The protocol first builds a taxonomy-filtered corpus, then generates contrastive multi-hop questions, and finally evaluates four selectors under the same candidate pool, evidence budget, generator, answer cap, and judge (Claude Opus 4.7, randomised answer-A/answer-B positions). The only intended source of variation is the evidence selector. We re-judge the six headline comparisons with a second strong proprietary judge from a different provider (DeepSeek V4 Pro, thinking mode) to assess inter-judge robustness (\secref{sec:inter-judge}).}
\label{fig:pipeline}
\end{figure*}

\section{Method}

We first define the selector, then the controls that make the comparisons interpretable. \figref{fig:pipeline} summarizes the full pipeline. The key rule is simple: every method sees the same candidates and uses the same generation settings. Only the evidence plan changes.

\subsection{GADMEC fitness}
GADMEC's fitness over a candidate evidence plan $\mathcal{P}$ combines five components:
\[
f(\mathcal{P}) = \alpha\,\textsc{cov} + \beta\,\textsc{div} + \gamma\,\textsc{cost} + \delta\,\textsc{coh} + \varepsilon\,\textsc{sub}
\]
Here \textsc{cov} rewards mean query similarity, and \textsc{div} rewards pairwise dissimilarity among selected chunks. \textsc{cost} is a normalised token penalty, \textsc{coh} is centroid--query similarity, and \textsc{sub} measures coverage of GPT-5.4-mini query sub-aspects embedded with all-MiniLM-L6-v2.

We use $\alpha = 0.30, \beta = 0.15, \gamma = 0.00, \delta = 0.15, \varepsilon = 0.40$, with sub-coverage threshold $0.40$. The BM25 lexical component $\zeta$ is $0$ for pure semantic GADMEC and activated only in the hybrid analyses.

\subsection{BRKGA decoder}
BRKGA \cite{goncalves2011brkga} represents each plan as random keys in $[0,1]^n$ and leaves feasibility to a decoder. The decoder sorts chunks by key and accepts them greedily until the plan reaches the 2000-token budget. It also enforces minimum query similarity $0.15$, redundancy threshold $0.80$, and at least three $k$-means thematic clusters. The search uses population size 20, elite fraction $0.24$, elite-inheritance probability $0.70$, at most 50 generations, early stopping after 15 stagnant generations, and seed $42$.

\subsection{Baselines}
We compare against greedy top-$k$ budget-fill, Maximal Marginal Relevance (MMR) with $\lambda=0.5$ \cite{carbonell1998mmr}, and the BM25 lexical retriever ($k_1=1.5, b=0.75$). All methods operate on the same top-$100$ cosine-pre-filtered candidate pool. Win-rate differences therefore reflect how methods choose evidence from the same pool, not whether one method had access to better candidates.

\subsection{Question generation and judging}
Question generation uses a contrastive multi-hop prompt with GPT-5.4-mini at $T=0.7$. Each area has 10 combinations, with 20 questions per combination. Answers use GPT-5.4-mini at $T=0$, seed $42$, evidence-only prompting, and \texttt{max\_completion\_tokens=300}. Claude Opus 4.7 judges randomised answer-A/answer-B pairs in a single pass using a fixed pairwise prompt. The prompt asks for a global preference based on factual correctness, completeness, evidence support, clarity, specificity of evidence-backed claims, multi-source synthesis, and coverage of the question aspects. Ties are allowed but reserved for answers that are truly equivalent; ties are recorded and excluded from the win-rate denominator.

\section{Experimental Setup}\label{sec:setup}

The setup applies the fixed-budget design to two scientific domains. Each pairwise judgment compares evidence selection under the same retrieval substrate, input budget, and answer-generation budget.

\subsection{Corpus}
The corpus is organised by a 3-level taxonomy: area $\to$ subarea $\to$ topic. We use two areas, Computer Science/Machine Learning (CS/ML; 341 papers) and Materials Science (MatSci; 346 papers), drawn from arXiv between 2024-01 and 2026-05. Each area contributes 10 cross-subfield combinations: five high-density TOP-regime combinations with at least 100 arXiv papers and five lower-density NICHO combinations with 10--50 papers. The manifest includes SHA-256 hashes over arXiv IDs.

\subsection{Fixed-budget control}
All methods use a 2000-token evidence budget with budget-fill. For compact tables and figures, we abbreviate GADMEC as GA. The resulting input lengths are tightly matched (median evidence tokens over 200 questions per area; parentheses give IQRs when shown):
\begin{center}
\footnotesize
\setlength{\tabcolsep}{4pt}
\begin{tabular}{lcc}
\toprule
Selector & CS/ML & MatSci \\
\midrule
GA & $1979$ ($1960$--$1993$) & $1973$ ($1938$--$1992$) \\
Greedy & $1986$ ($1966$--$1994$) & $1979$ ($1960$--$1992$) \\
MMR & $1984$ & $1982$ \\
BM25 & $1986$ ($1972$--$1995$) & $1987$ ($1967$--$1994$) \\
\bottomrule
\end{tabular}
\end{center}
All methods fill approximately $99$\,\% of the budget. Answer generation also uses identical caps and settings. Thus systematic answer-length differences are not caused by unequal budgets; they reflect what the selected evidence makes available to the generator.

\subsection{Pre-registration}
The primary registration locks four hypotheses of the form $\mathrm{WR}(\mathrm{GA}\ \text{vs.}\ \textsc{base})>0.5$, where WR denotes win rate. Each hypothesis uses one-sided cluster bootstrap over combination identifiers (10\,000 resamples) and Bonferroni $\alpha=0.0125$. The addenda lock the $\leq200$-character length-matched bin, ablation predictions, BM25 parity checks, and hybrid predictions before judging. The deviation log records analyses added later, including bin sensitivity, content-distance slicing, and hybrid configurations. It also records two drift items: Opus 4.7 rather than the pre-registered 4.6 judge, and 200 rather than 100 questions per area.

\section{Results}

\subsection{Main results}

We report results in four blocks. Block 1 gives the pre-registered semantic comparison against Greedy and MMR. Block 2 adds BM25 as a lexical baseline under the same budget. Block 3 tests whether the BRKGA machinery alone explains the gains. Block 4 asks whether adding lexical signal to the GADMEC fitness recovers the BM25 gap. The order is diagnostic: each block removes a different explanation for the observed win rates.

\textbf{Block 1 --- Primary (pre-registered, Bonferroni $\alpha = 0.0125$ over 4 tests).}
\tabref{tab:primary} reports the four primary semantic comparisons.

\begin{table}[ht]
\centering
\caption{Primary semantic comparisons.}
\label{tab:primary}
\scriptsize
\setlength{\tabcolsep}{3pt}
\begin{tabular}{@{}llrrrrr@{}}
\toprule
Comparison & Area & WR \% & $p^{\mathrm{V}}$ & $p^{\mathrm{W}}$ & $p^{\mathrm{P}}$ & Bonf \\
\midrule
GADMEC vs Greedy & CS/ML  & 57.4 & 0.0251 & 0.0515 & 0.0586 & $\times$ \\
GADMEC vs Greedy & MatSci & 58.9 & 0.0205 & 0.0394 & 0.0391 & $\bullet$ \\
GADMEC vs MMR    & CS/ML  & 60.1 & 0.0000 & \textbf{0.0043} & \textbf{0.0078} & $\star$ \\
GADMEC vs MMR    & MatSci & 61.7 & 0.0025 & 0.0201 & 0.0234 & $\bullet$ \\
\bottomrule
\end{tabular}
\end{table}

We report three p-values, each with a separate role. $p^{\mathrm{V}}$ comes from a vanilla cluster bootstrap by combination identifier (10,000 resamples). $p^{\mathrm{W}}$ comes from a pivotal wild-cluster bootstrap with Webb 6-point weights \cite{cameron2008boostrap}, centred at the null $H_0\!:\!\mathrm{WR}=0.5$. $p^{\mathrm{P}}$ comes from an exact cluster sign-flip permutation test. Because each area has 10 clusters, the exact test can enumerate all $2^{10}=1024$ sign assignments. It therefore does not rely on large-cluster asymptotics.

The wild bootstrap and the exact sign-flip test agree to within $0.005$ on every primary comparison. This makes the small-cluster concern visible rather than hidden in a single number. For the four-test primary family, Bonferroni decisions use $p^{\mathrm{W}}$ at $\alpha = 0.0125$: $\star$ passes Bonferroni, $\bullet$ passes uncorrected $\alpha=0.05$, and $\times$ fails. One of four primary tests passes Bonferroni (GADMEC vs MMR in CS/ML); two more pass only before correction. The gap between $p^{\mathrm{V}}$ and the cluster-aware p-values reflects per-cluster heterogeneity (cross-combination standard deviation of win rate $\approx 12$ percentage points; see \texttt{results/all\_bootstrap\_pvalues.json}). We return to this point in \S6.

\textbf{Block 2 --- Secondary (BM25 retrieval baseline, addendum, descriptive).}
\tabref{tab:bm25} adds the lexical BM25 baseline under the same budget.

\begin{table}[ht]
\centering
\caption{BM25 lexical-baseline comparisons.}
\label{tab:bm25}
\scriptsize
\setlength{\tabcolsep}{3pt}
\begin{tabular}{@{}llrrr@{}}
\toprule
Comparison & Area & WR \% & $p^{\mathrm{W}}$ & Bonf \\
\midrule
GADMEC vs BM25 & CS/ML  & 40.2 & 0.0188 & $\bullet$ \\
GADMEC vs BM25 & MatSci & 39.2 & \textbf{0.0007} & $\star$ \\
\bottomrule
\end{tabular}
\end{table}

BM25 beats pure semantic GADMEC in both domains under the same budget. The effect is Bonferroni-significant in MatSci ($p^{\mathrm{W}}=0.0007$) and significant at uncorrected $\alpha=0.05$ in CS/ML ($p^{\mathrm{W}}=0.0188$). Dense pre-filtering therefore does not subsume lexical signal. Block 4 tests whether adding that lexical signal to GADMEC's fitness recovers the gap.

\textbf{Block 3 --- Ablations (descriptive, addendum-locked).}
\tabref{tab:ablations} tests whether the BRKGA machinery alone explains the gains.

\begin{table}[ht]
\centering
\caption{Ablations against Greedy.}
\label{tab:ablations}
\scriptsize
\setlength{\tabcolsep}{3pt}
\begin{tabular}{@{}lcc@{}}
\toprule
Ablation vs Greedy & CS/ML WR & MatSci WR \\
\midrule
random\_fitness                  & 42.9 [36.7, 49.2] & 37.2 [29.6, 43.2] \\
random\_fitness + no\_constr.    & 39.8 [34.7, 45.9] & 36.7 [29.9, 43.0] \\
no\_subaspect ($\varepsilon{=}0$) & 57.4 [51.0, 63.5] & 55.8 [49.2, 62.8] \\
\bottomrule
\end{tabular}
\end{table}

\begin{figure}[t]
\centering
\includegraphics[width=\columnwidth]{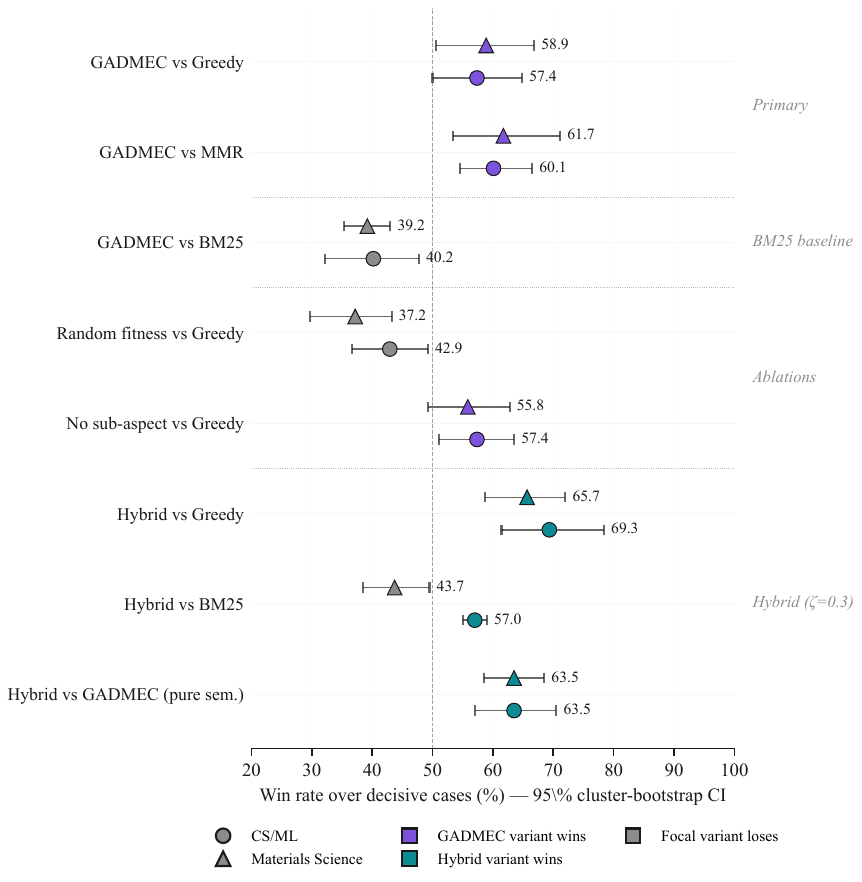}
\caption{Main results. Each row is one pairwise comparison. Markers: circles = CS/ML, triangles = Materials Science. Colours: Slate Blue = GADMEC variant wins, Dark Teal = hybrid variant wins, grey = focal variant loses. Horizontal bars are 95\% cluster-bootstrap confidence intervals (CIs) by combination identifier; dashed vertical line is the 50\% no-effect reference.}
\label{fig:main}
\end{figure}

\textbf{Block 4 --- Hybrid GADMEC ($\zeta = 0.3$, exploratory, locked before judging).}

Block 2 shows that lexical signal matters. We therefore activate the BM25 component of the surrogate fitness ($\zeta = 0.3$) while leaving the other fitness weights unchanged. For each question, we compare the hybrid against Greedy, BM25, and original semantic GA ($\zeta = 0$). \tabref{tab:hybrid} reports the hybrid comparisons:

\begin{table}[ht]
\centering
\caption{Hybrid GADMEC comparisons.}
\label{tab:hybrid}
\scriptsize
\setlength{\tabcolsep}{3pt}
\begin{tabular}{@{}llrrr@{}}
\toprule
Hybrid vs & Area & WR \% & $p^{\mathrm{W}}$ & Bonf \\
\midrule
Greedy               & CS/ML  & 69.3 & \textbf{0.0016} & $\star$ \\
Greedy               & MatSci & 65.7 & \textbf{0.0021} & $\star$ \\
BM25                 & CS/ML  & 57.0 & \textbf{0.0007} & $\star$ \\
BM25                 & MatSci & 43.7 & 0.0335 & $\bullet$ \\
Pure semantic GADMEC & CS/ML  & 63.5 & \textbf{0.0012} & $\star$ \\
Pure semantic GADMEC & MatSci & 63.5 & \textbf{0.0015} & $\star$ \\
\bottomrule
\end{tabular}
\end{table}

Block 4 is a separate family of six comparisons. With within-family Bonferroni correction ($\alpha = 0.05 / 6 \approx 0.0083$), five of six hybrid comparisons pass under wild-cluster inference. The exception is hybrid vs BM25 in MatSci: it passes at uncorrected $\alpha = 0.05$, but not Bonferroni, and it remains directionally a \emph{loss} (win rate $43.7\%$).

Taken together, the hybrid reverses the BM25 deficit in CS/ML (hybrid vs BM25 $p^{\mathrm{W}}=0.0007$), narrows but does not eliminate the BM25 deficit in MatSci ($-10.8$ percentage points before, $-6.3$ after), and improves over both semantic baselines and pure semantic GADMEC. Pre-registered predictions for $\zeta = 0.3$ (deviations entry 7, locked before judging) were vs Greedy $[55, 65]\%$, vs BM25 $[45, 60]\%$, and vs pure semantic GADMEC $[50, 60]\%$. Observed values met or exceeded the predicted interval in 5 of 6 cells; the exception was hybrid vs BM25 in MatSci ($43.7\%$), marginally below the $[45, 60]$ lower bound. \figref{fig:main} summarizes Blocks 1--4 with confidence intervals.

\subsection{Length-matched robustness}

Length matching is a mechanism check, not a second primary endpoint. If the advantage is mostly a length artifact, it should weaken when GADMEC and Greedy answers have similar character counts. Bracketed intervals are 95\% confidence intervals (CIs). \tabref{tab:length-matched} reports the pre-registered $\leq 200$-character matched bin:

\begin{table}[ht]
\centering
\caption{Length-matched robustness for GADMEC vs Greedy.}
\label{tab:length-matched}
\scriptsize
\setlength{\tabcolsep}{3pt}
\begin{tabular}{@{}lcc@{}}
\toprule
Subset & CS/ML WR [CI] & MatSci WR [CI] \\
\midrule
All pairs              & 57.4 [50.0, 64.8] & 58.9 [50.5, 66.8] \\
\textbf{In matched bin} & \textbf{50.0 [39.8, 60.4]} & \textbf{56.1 [46.4, 65.2]} \\
Out of matched bin     & 64.4 [56.9, 71.6] & 62.2 [51.0, 73.8] \\
\bottomrule
\end{tabular}
\end{table}

\begin{figure}[t]
\centering
\includegraphics[width=\columnwidth]{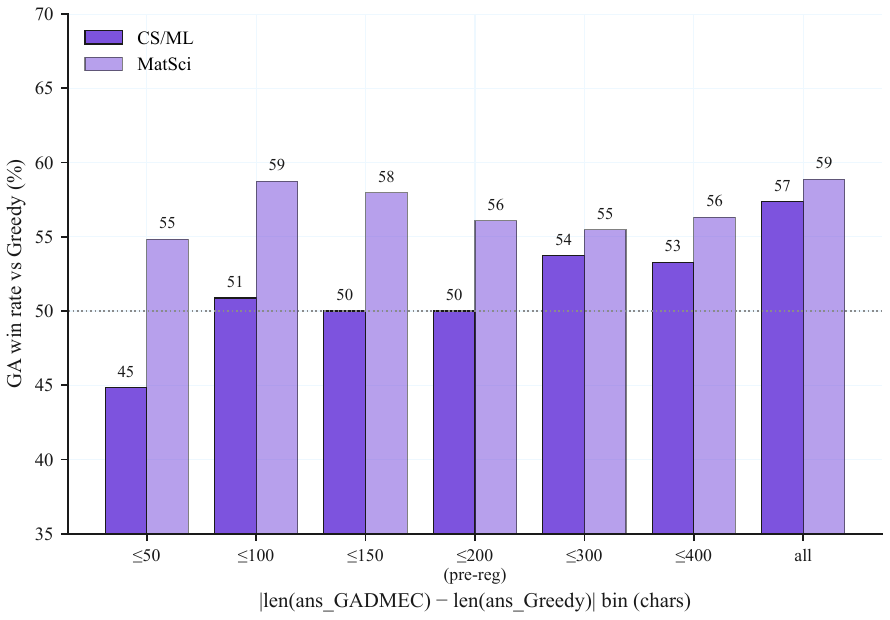}
\caption{Bin sensitivity. The figure varies the allowed answer-length difference between GADMEC and Greedy. MatSci remains stable across bin widths, while CS/ML moves toward parity as the bin tightens; this is consistent with a less length-entangled MatSci advantage.}
\label{fig:bins}
\end{figure}

We then vary the bin width as an exploratory sensitivity sweep (\tabref{tab:bin-sensitivity}; \figref{fig:bins}).

\begin{table}[ht]
\centering
\caption{Bin-width sensitivity for length matching.}
\label{tab:bin-sensitivity}
\scriptsize
\setlength{\tabcolsep}{4pt}
\begin{tabular}{@{}lcc@{}}
\toprule
Bin width & CS/ML WR \% & MatSci WR \% \\
\midrule
$\leq 50$            & 44.8 & 54.8 \\
$\leq 100$           & 50.9 & 58.7 \\
$\leq 150$           & 50.0 & 58.0 \\
$\leq 200$ (pre-reg) & 50.0 & 56.1 \\
$\leq 300$           & 53.7 & 55.5 \\
$\leq 400$           & 53.3 & 56.3 \\
all                  & 57.4 & 58.9 \\
\bottomrule
\end{tabular}
\end{table}

MatSci remains in the 54--58\,\% range across bin widths. CS/ML moves toward parity and drops to 44.8\,\% at the tightest bin, where most answers are too short to differ structurally. This domain split is exploratory rather than confirmatory. The pre-registered $\leq 200$-character bin is not Bonferroni-significant in either area, so the stable MatSci range should be read as a sensitivity pattern for follow-up work, not as a settled domain difference.

\subsection{Content-distance slicing within matched bin (exploratory)}

Within the matched bin, we ask whether answer pairs that differ more in content show a stronger GADMEC signal. \tabref{tab:jaccard-slices} slices pairs by quartiles of Jaccard similarity between GA and Greedy answers:

\begin{table}[ht]
\centering
\caption{Jaccard slices within the matched bin.}
\label{tab:jaccard-slices}
\scriptsize
\setlength{\tabcolsep}{4pt}
\begin{tabular}{@{}lcc@{}}
\toprule
Jaccard slice & CS/ML WR \% & MatSci WR \% \\
\midrule
Q1 (most distinct) & 62.5 ($n{=}24$) & 65.4 ($n{=}26$) \\
Q2                 & 45.8 & 55.6 \\
Q3                 & 66.7 & 44.4 \\
Q4 (most similar)  & 25.0 & 59.3 \\
\bottomrule
\end{tabular}
\end{table}

\begin{figure}[t]
\centering
\includegraphics[width=\columnwidth]{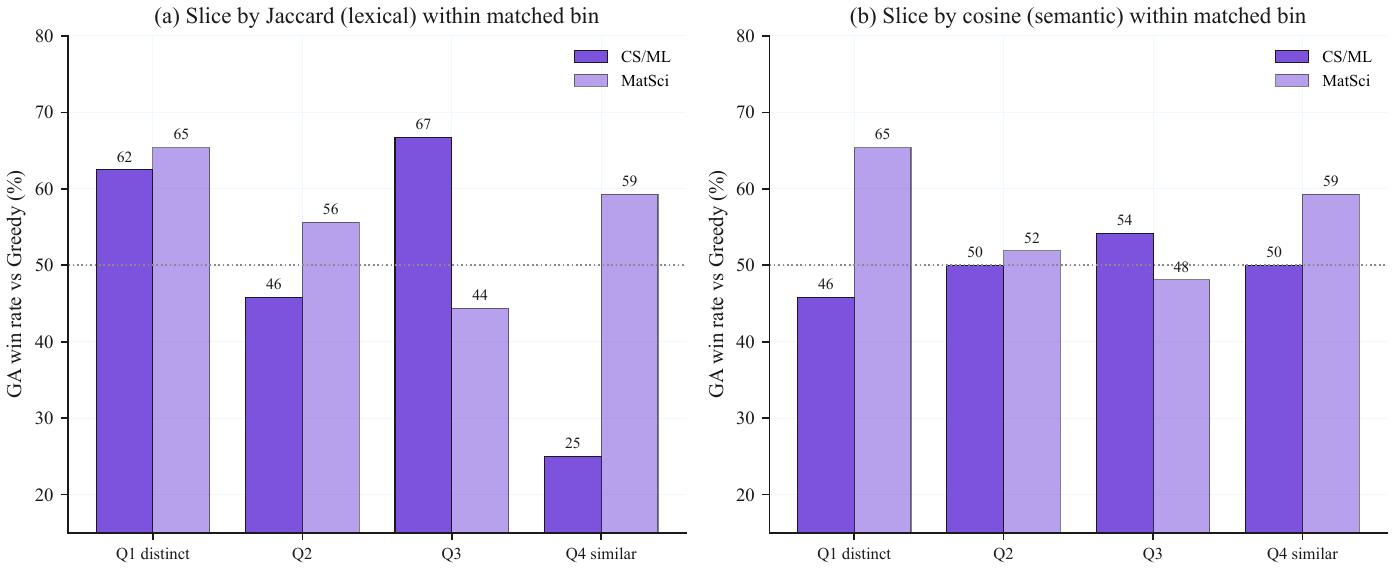}
\caption{Content-distance slicing within the matched bin. Panel (a) groups answer pairs by lexical overlap; panel (b) groups them by semantic cosine similarity. MatSci is stable in the most-distinct bucket under both definitions, while CS/ML depends on the distance measure.}
\label{fig:slices}
\end{figure}

Across four additional slicing specifications, MatSci's most-distinct bucket stays at 60--66\,\%. CS/ML is specification-dependent: lexical Jaccard shows a 62.5\,\% Q1 win rate, but cosine Q1 drops to 45.8\,\%. One plausible reason is geometric: CS/ML answers occupy a narrower embedding-space region (cosine range 0.67--0.97) than MatSci (0.53--1.00), so cosine slices discriminate less in CS/ML. \figref{fig:slices} compares the lexical and semantic-distance slices.

\subsection{Ablation discussion}

\paragraph{Fitness signal contribution.} Randomising the surrogate fitness while keeping the decoder machinery drops win rate to 42.9\,\% in CS/ML and 37.2\,\% in MatSci. Disabling constraints as well lowers performance slightly further (39.8\,\%, 36.7\,\%). Thus the fitness signal, not merely the BRKGA decoder, accounts for the 14.5--21.7 percentage-point gap to full GADMEC.

\paragraph{Sub-aspect coverage.} Removing the $\varepsilon$ sub-aspect component is inert in CS/ML (57.4\,\% vs.\ 57.4\,\%) and costs 3.1 percentage points in MatSci (58.9\,\% to 55.8\,\%). This is a domain finding rather than a calibration failure: MatSci appears to benefit from explicit decomposable-query coverage, while CS/ML's remaining gain is more entangled with length and lexical content distance.

\subsection{Inter-judge agreement}\label{sec:inter-judge}

We re-judge the four primary tests and the two hybrid-vs-BM25 comparisons with DeepSeek V4 Pro (thinking mode), a second strong proprietary judge from a different provider. We use the same prompt, answer-A/answer-B randomisation, and tie-handling rule; ties are recorded but excluded from the win-rate denominator. MatSci primary uses $n=199$ shared pairs (one Opus pair was a parse-error retry; Opus WR is recomputed on the matched subset for this section). The two judges come from different providers (Anthropic and DeepSeek). \tabref{tab:inter-judge} reports Cohen's $\kappa$ and Gwet's agreement coefficient (AC1):

\begin{table}[ht]
\centering
\caption{Inter-judge agreement.}
\label{tab:inter-judge}
\scriptsize
\setlength{\tabcolsep}{3pt}
\begin{tabular}{@{}lrrrrr@{}}
\toprule
Comparison & Opus WR & DeepSeek WR & raw \% & $\kappa$ & AC1 \\
\midrule
CS/ML GA vs Greedy        & 57.4 & 57.4 & 87.0 & 0.745 & 0.826 \\
CS/ML GA vs MMR           & 60.1 & 62.2 & 85.5 & 0.705 & 0.808 \\
MatSci GA vs Greedy       & 58.7 & 61.1 & 84.9 & 0.693 & 0.800 \\
MatSci GA vs MMR          & 62.1 & 59.4 & 86.4 & 0.731 & 0.819 \\
CS/ML hybrid vs BM25      & 57.0 & 52.0 & 83.0 & 0.658 & 0.663 \\
MatSci hybrid vs BM25     & 43.7 & 41.7 & 77.0 & 0.535 & 0.694 \\
\bottomrule
\end{tabular}
\end{table}

Cohen's $\kappa$ ranges from moderate (0.535, MatSci hybrid vs BM25) to substantial (0.745, CS/ML GA vs Greedy) per \citet{landis1977measurement}. Gwet's AC1 is in the substantial-to-almost-perfect range (0.663--0.826). Win-rate point estimates differ by at most 5 percentage points, and the directional verdict matches in all six comparisons. This reduces the risk that the headline depends on one judge model. It does not replace human gold annotation, which we leave to follow-up work.

\section{Discussion}

The results are best read as a mechanism map, not as a ranking of retrievers. Fixed budgets remove one structural confound. What remains separates into three signals: semantic composition, lexical matching, and answer-length effects.

\paragraph{Composition vs.\ length under fixed budgets.} Even after matching input budgets and answer caps, a method can win in several ways. Its evidence may induce longer answers. It may support denser answers of similar length. Or the judge may still reward small length differences. The domain split suggests that these mechanisms should not be collapsed into a single ``better retrieval'' story.

\paragraph{MatSci: composition-driven (exploratory).} Across seven bin widths and five content-distance specifications, GADMEC's matched-bin advantage is stable (54--66\,\%). In the pre-registered matched bin, GADMEC is slightly longer more often than Greedy ($63$ vs.\ $44$ cases at $\leq200$ chars). This suggests that composition can show up as length in MatSci, rather than length simply causing the judged advantage. The reading remains exploratory: the MatSci CI in the pre-registered bin is $[46.4, 65.2]$, which crosses $50\%$.

\paragraph{CS/ML: composition entangled with length.} CS/ML behaves differently. The composition signal appears under lexical content-distance (Q1 Jaccard 62.5\,\%) but not under semantic distance (Q1 cosine 45.8\,\%). CS/ML answers cluster in a narrower embedding region, so semantic-distance slicing has lower discriminating power. The mechanism is more entangled with length in CS/ML than in MatSci.

\paragraph{BM25 as complementary signal.} BM25's dominance over pure semantic GADMEC shows that lexical signal still matters after dense pre-filtering. The hybrid beats BM25 in CS/ML and narrows, but does not close, the MatSci gap. We therefore treat lexical matching and semantic composition as complementary signals.

\paragraph{Inference under cluster heterogeneity.} The gap between $p^{\mathrm{V}}$ and $p^{\mathrm{W}}$ in Block 1 is not a quirk. It is what happens when cluster correlation is taken seriously in a small-cluster design. Dual reporting of vanilla and wild-cluster bootstrap is standard in econometrics, but rare in NLP evaluation. We recommend it for cluster-structured NLP benchmarks where some categories, domains, or sources may be easier than others. The test choice can change the headline, so the cost of that choice should be visible to the reader.

\paragraph{Scope of the proposed standard.} The cluster issue is not specific to this benchmark. Many LLM-as-a-judge evaluations have natural clusters. MT-Bench groups 80 multi-turn questions into eight capability categories; AlpacaEval draws 805 instructions from five source datasets; MMLU spans 57 subjects in four broad domains; and MTEB spans eight task families across 58 datasets. Multi-hop RAG benchmarks such as HotpotQA and 2WikiMultiHop also build questions around bridge or comparison entities. In each case, some clusters may be easier, harder, or more favorable to one method. A binomial p-value over pooled pairs can therefore overstate confidence. We do not test the standard on those benchmarks here, but the recommendation is meant for cluster-structured pairwise judging designs in general.

\section{Conclusion}

Fixed-budget evaluation makes multi-hop RAG comparisons easier to interpret. Under matched evidence budgets and answer caps, pure semantic GADMEC beats Greedy and MMR directionally in all four pre-registered tests. Under wild-cluster bootstrap, however, only one test passes Bonferroni and two more pass only at uncorrected $\alpha = 0.05$. A standard binomial test would have reported all four as significant.

The same protocol also changes how we read retrieval signals. BM25 beats pure semantic GADMEC in both domains, showing that lexical matching remains important under the same budget. A lexical-semantic hybrid reverses the BM25 gap in CS/ML and narrows it in MatSci without closing it. Length-matched analyses suggest a domain split. This claim remains exploratory: in the pre-registered $\leq 200$-character bin neither area reaches Bonferroni significance and the MatSci CI $[46.4, 65.2]$ crosses $50\%$. Code, data manifests, registrations, deviations, prompts, and judgments are available in the project repository. \appref{app:heterogeneity}, \appref{app:cost}, and \appref{app:prereg} provide descriptive diagnostics, cost logs, and registration summaries.

\section{Limitations}

The conclusions are bounded by the evaluation instrument. The main limits are the judge, the number of domains, the exploratory status of several diagnostics, the small number of clusters, and the current granularity of the mechanism analysis.

\paragraph{Judge robustness.} The primary judge is Claude Opus 4.7. We replicate the six pairwise comparisons that drive the headline (4 primary + 2 hybrid vs BM25) with DeepSeek V4 Pro in thinking mode, a second strong proprietary judge from a different provider. Agreement is moderate-to-substantial across all six comparisons. Cohen's $\kappa \in [0.535, 0.745]$ \cite{landis1977measurement}, Gwet's AC1 $\in [0.663, 0.826]$, raw agreement is $77$--$87\%$, and the directional verdict matches in all six (\secref{sec:inter-judge}). Win-rate point estimates differ by at most 5 percentage points. We did not replicate the BM25-only and ablation comparisons; full multi-judge replication is future work.

\paragraph{Two domains.} The domain-dependent mechanism is based on CS/ML and MatSci only. Replication on biomedical, legal, or social-science corpora is needed.

\paragraph{Exploratory analyses.} The bin-sensitivity sweep, content-distance slicing, hybrid $\zeta=0.3$, and MatSci $\zeta=0.5$ analyses were added after the primary results, although their predictions were locked before judging where applicable. Primary inference rests on the pre-registered tests. The MatSci matched-bin CI $[46.4, 65.2]$ crosses $50\%$, so the strongest MatSci robustness claim should be read as an exploratory pattern.

\paragraph{Cluster count.} We have 10 combinations per area, which is at the lower edge of what wild-cluster bootstrap is recommended for ($n_{\mathrm{clusters}} \geq 20$--30 in \citealp{cameron2008boostrap}). We address this by also reporting the exact cluster sign-flip permutation test. It enumerates all $2^{10}=1024$ sign assignments and is unaffected by the small cluster count. The two tests agree to within $0.005$ on every primary comparison (and within $0.01$ across all reported comparisons), so the Block 1 inference is not driven by wild-bootstrap asymptotics alone. Replication on benchmarks with more cluster-level diversity would improve external validity and precision.

\paragraph{Mechanistic granularity.} We do not yet quantify which chunks GADMEC selects differently from Greedy. Chunk-overlap and evidence-structure analyses are needed to isolate the CS/ML residual mechanism.

\paragraph{Necessity of each standard component.} The proposed measurement standard has four components. This paper directly tests two: fixed-budget design (controlled in the experimental setup; \secref{sec:setup}) and cluster-aware inference (Block 1 dual reporting plus the permutation check). Pre-registration with a deviation log and second-judge replication constrain reporting, but they are not variables we can ablate after the fact. We therefore do not claim that removing pre-registration would change conclusions in this dataset. Future work should test each component across multiple benchmarks.

\section*{Acknowledgments}

Camilo Chacón Sartori acknowledges funding from R\&D Project PID2022-138283NBI00, funded by MICIU/AEI/10.13039/501100011033 and ``FEDER -- A way of making Europe''.

\bibliography{custom}

\clearpage
\appendix

\section{Supplementary Diagnostics}
\label{app:heterogeneity}

The appendix is descriptive. It does not introduce new primary claims; instead, it makes the aggregate results easier to audit. \figref{fig:combo-heatmap}, \figref{fig:joint-outcome}, and \figref{fig:regime} answer three diagnostic questions: whether the aggregate win rates are driven by a few combinations, whether wins against Greedy and MMR tend to co-occur, and whether the TOP/NICHO corpus regime changes the interpretation.

\tabref{tab:qual-examples} adds three logged examples to make the mechanisms less abstract. These examples are illustrative only: they are selected from existing answer-pair and verdict logs, and they do not define a new hypothesis test.

\begin{center}
\scriptsize
\setlength{\tabcolsep}{2pt}
\begin{tabular}{@{}p{0.24\columnwidth}p{0.36\columnwidth}p{0.32\columnwidth}@{}}
\toprule
Logged example & Preferred answer adds & Mechanism illustrated \\
\midrule
CS/ML, GADMEC $>$ Greedy, $q{=}3$ & Covers reward/Kullback--Leibler (KL) infringement control and spatial-aware compositional generation; Greedy supports only the infringement side. & Semantic composition improves sub-aspect coverage under the same budget. \\
CS/ML, BM25 $>$ GADMEC, $q{=}2$ & Gives details for the two-stage structural pipeline and entity-relation mining; pure semantic GADMEC is correct but terser. & Lexical matching recovers method and control terms missed by embedding-only fitness. \\
CS/ML, Hybrid $>$ BM25, $q{=}4$ & Retrieves constraint-feedback and semantic action/entity control evidence; BM25 lacks the law-grounded comparison. & Lexical signal inside semantic fitness can recover part of the BM25 gap. \\
\bottomrule
\end{tabular}
\captionsetup{type=table,hypcap=false}
\caption{Illustrative examples from logged answer pairs and judge verdicts. The table is descriptive and is not used for statistical inference. It shows how the aggregate mechanisms in the main text appear in individual comparisons.}
\label{tab:qual-examples}
\end{center}

\subsection{Per-combination win rates}
\begin{figure*}[t]
\centering
\includegraphics[width=\textwidth]{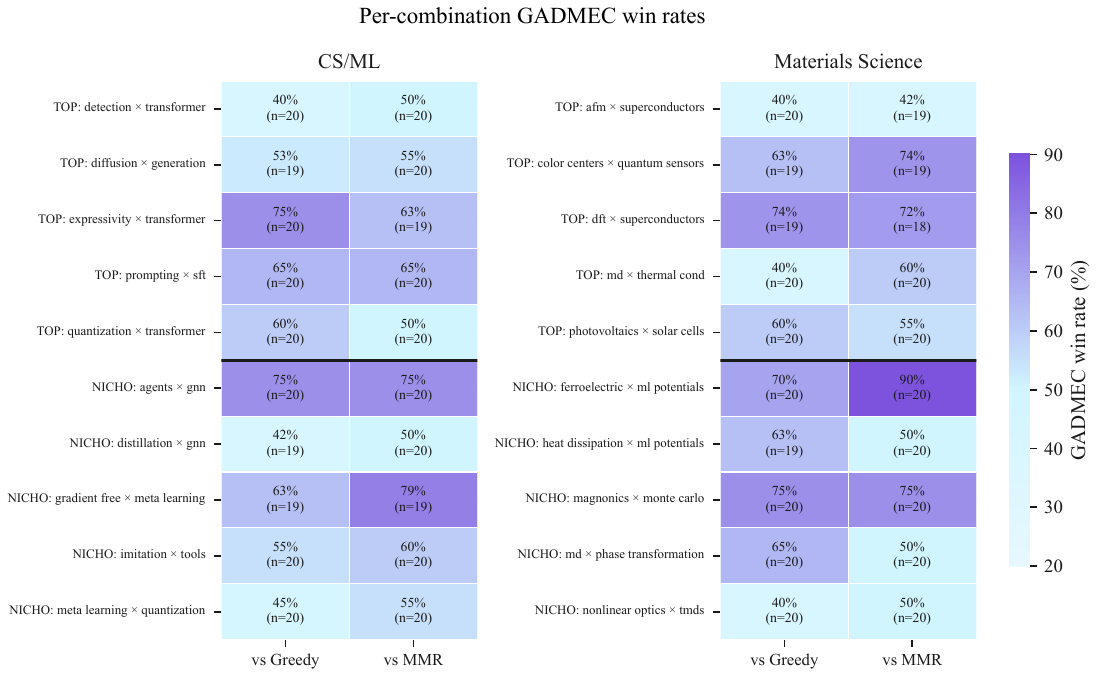}
\caption{Pure semantic GADMEC win rate per combination, stratified by domain and corpus regime (TOP/NICHO). Each panel is one domain; each row is a cross-subfield combination; the two columns report win rates against Greedy and MMR. The heatmap is a descriptive stability check: it exposes heterogeneity behind the aggregate rates without adding a separate hypothesis test.}
\label{fig:combo-heatmap}
\end{figure*}

\figref{fig:combo-heatmap} should be read as a stability check, not as a separate hypothesis test. The main text reports cluster-bootstrap intervals over combination identifiers; this figure shows the same source of variation directly. It also helps identify where a future mechanistic analysis should focus: combinations with high semantic-GADMEC win rates likely contain evidence sets where coverage and diversity help, while low-rate combinations are candidates for lexical or length-driven failure modes.

\subsection{Joint outcome matrix}
\begin{figure}[ht]
\centering
\includegraphics[width=0.95\columnwidth]{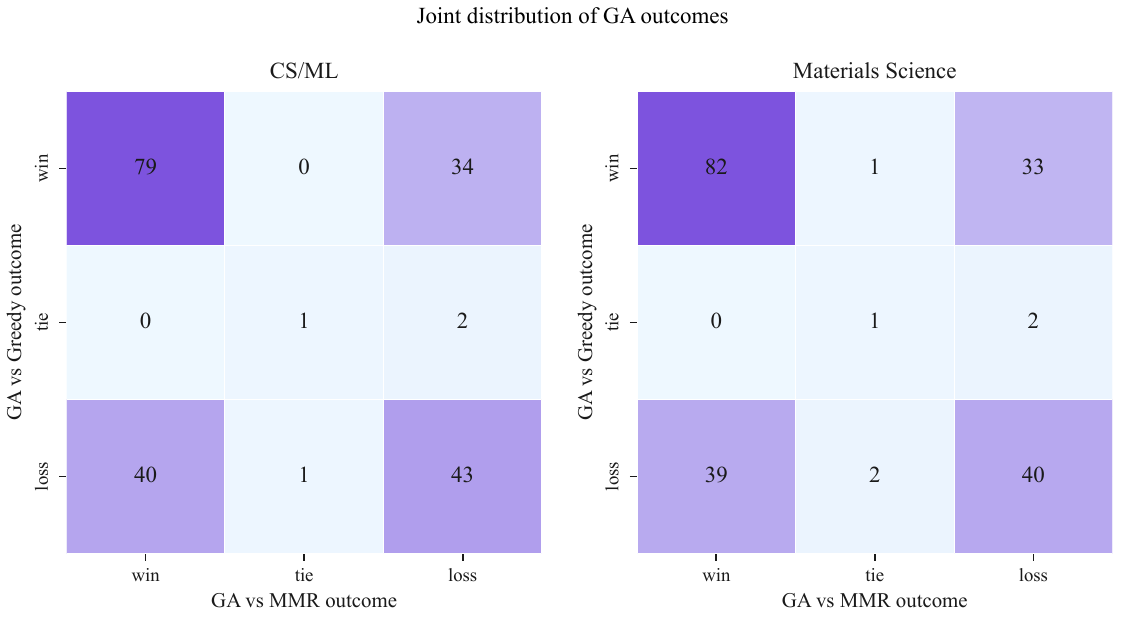}
\caption{Joint outcome matrix for GADMEC against Greedy and MMR, by area. Rows give verdicts against Greedy and columns give verdicts against MMR. Off-diagonal cells show where the two semantic baselines expose different failure modes.}
\label{fig:joint-outcome}
\end{figure}

\figref{fig:joint-outcome} gives a more concrete view of what the aggregate primary tests mean. In both domains, many questions are wins against both semantic baselines, but there are also substantial off-diagonal cells. Thus Greedy and MMR are not redundant controls: they overlap, but they do not define the same comparison.

\subsection{Stratified TOP/NICHO analysis}
\begin{figure}[ht]
\centering
\includegraphics[width=0.95\columnwidth]{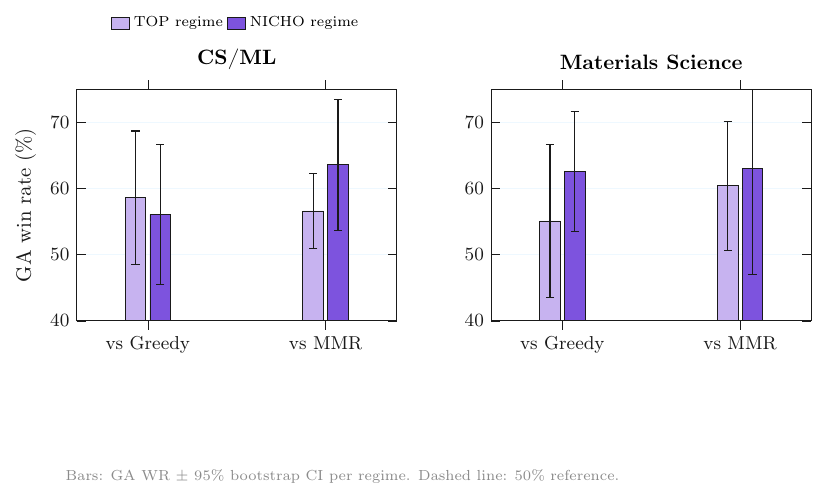}
\caption{Pure semantic GADMEC win rates stratified by corpus regime. TOP denotes dense topic pairs; NICHO denotes lower-density pairs. Bars show win rates and 95\% bootstrap intervals within each regime.}
\label{fig:regime}
\end{figure}

\figref{fig:regime} is included because the corpus construction deliberately mixes common and niche cross-subfield combinations. The figure is reassuring only in a limited sense: it shows no obvious collapse in either regime, but the intervals are wide enough that the main paper should rely on the pooled, pre-registered tests rather than regime-specific claims.

\section{Cost log}
\label{app:cost}
Total judging cost across all conditions (primary + BM25 + 3 ablations + hybrid $\zeta=0.3$ + hybrid $\zeta=0.5$ MatSci): approximately \texteuro{}82 (\$95 USD, May 23 2026 European Central Bank rate $0.8605$ EUR/USD), via the Claude Opus 4.7 application programming interface (API), billed at \$15/1M input tokens and \$75/1M output tokens. Per-run logs are included in \texttt{results/cost\_v3\_judging.json} and \texttt{results/cost\_v3\_extras\_judging.json}.

The cost log is reported for audit, not as a metric. It makes the evaluation scale visible: every reported comparison uses the same evidence budget, generator, answer cap, judge, and answer-position randomisation, so the paired design carries the inference rather than any cost normalisation.

\section{Pre-registration documents}
\label{app:prereg}
The prepared release archive includes the primary pre-registration, dated 2026-05-19; the addendum, dated 2026-05-23; and the running deviation log.

The primary registration fixes the four pairwise hypotheses, the Bonferroni correction, tie handling, and cluster-bootstrap reporting before judging. The addendum records analyses that became necessary after inspecting the design, including length matching and ablations. The deviation log separates later descriptive checks, such as bin-width sensitivity, content-distance slicing, and hybrid lexical weights.

This structure is meant to keep the paper readable without blurring evidential status. The main claims come from registered fixed-budget comparisons. The remaining analyses explain where the signal is stable, where it is domain-dependent, and which mechanisms should be tested in follow-up work rather than treated as settled here.

\end{document}